# CODE SWITCHING LANGUAGE MODEL USING MONOLINGUAL TRAINING DATA

*Asadullah, Tauseef Ahmed*

asadullah73@ce.ceme.edu.pk, tauseef.ahmed@ttu.ee

**ABSTRACT**

Training a code-switching (CS) language model using only monolingual data is still an ongoing research problem. In this paper, a CS language model is trained using only monolingual training data. As recurrent neural network (RNN) models are best suited for predicting sequential data. In this work, an RNN language model is trained using alternate batches from only monolingual English and Spanish data and the perplexity of the language model is computed. From the results, it is concluded that using alternate batches of monolingual data in training reduced the perplexity of a CS language model. The results were consistently improved using mean square error (MSE) in the output embeddings of RNN based language model. By combining both methods, perplexity is reduced from 299.63 to 80.38. The proposed methods were comparable to the language model fine tune with code-switch training data.

## 1. RELATED WORK

Recurrent neural networks (RNNs) have outperformed language model nowadays as compared to classical language model [1]. RNN language model has also improved the performance of automatic speech recognition (ASR) system [2]. The main reason is that RNN model works well when the input is a sequence of data [3]. However, code-switching (CS) language model is still an ongoing research problem [4]. CS is a phenomenon in which more than one language occurs within a document. CS can occur across sentences or within a sentence. Usually, a speaker is speaking native language but sometimes switch to other language. The dominant language is called matrix language while the submissive language is called embedding matrix. The CS phenomena usually happened in a diverse and multi-cultural society.

Related research is focused on monolingual training data where code-switching training data is hard to collect. However, monolingual data for individual languages are available in excess. The performance of the language model will not degrade if we have code-switch training data. The real challenge is that when we have monolingual training data and code-switch test data.

There has also been competitions and challenges for building a code-switched speech technology e.g. Speech technologies for code-switching in Multi-lingual communities by Microsoft [5].

Current research is focused on only monolingual training data. Recent work [6] integrates two monolingual language models. Other works [7, 8] generates synthetic data from the model distribution of data. They then used a switch to detect language model and incorporate the detected language model in automatic speech recognition. Chuang et. all [9] used only monolingual data to train a CS language model. They minimized Kullback-Leibler (KL) divergence between output embedding matrices of English and Chinese monolingual data during training. They consistently improved the results by normalizing output embedding matrices. Similar work has been found in [10] where they trained a CS speech recognition system by minimizing Jensen-Shannon (JS) divergence between output embeddings of end-to-end speech recognition during training. The work done in [11] has used monolingual data in training and then fine-tuned with small code-switch training data. They also synthesized code-switch data in order to train a speech recognition using discriminative method. Previous work [12] focused on different combinations of monolingual and code-switching training data and identify and evaluate a CS language model. More recent works [13, 14] are focused on meta-transfer learning approach where they improved a code-switching end-to-end speech recognition. The advantage of meta-transfer learning is that the model learns both monolingual data and cod-switched data.

The work presented in this paper is based on a typical RNN language model. RNN model is usually used for predicting sequential data, but here in this case, alternate batches of monolingual data (English, Spanish) are used during RNN model training. This approach is used to reduce the embedding space of both languages and minimize the perplexity of the code-switched test data. Furthermore, mean square error (MSE) between the output embeddings of RNN network is used for regularization between the output embeddings of both languages in training.

Rest of the paper is organized as follows. Section 2 covers methods used in this research. Section 3 explains our experiments including dataset, implementation and results. Finally, the section 4 presents the conclusion of this research work.

## 2. METHODS

In this work, the RNN language model is used as research tool[1]. The RNN is used to predict next word given a sequence of words history. Technically, it takes a non-linear transformation ($W$), hidden state ($h_i$) and a final layer softmax to compute the probability of next word. Mathematically, it can be represented by:

$$y_i = \text{softmax}(W\ h_i) \quad \ldots \ldots \ldots \ldots \ldots \ldots (1)$$

where $y_i$ is the prediction word, W is a transformation matrix and $h_i$ is the RNN hidden state. Specifically, $W \in \mathbb{R}^{V \times z}$ where $V$ is the vocabulary size and z is the hidden dimension. Usually, RNN model predict next word of a sequence of words in a monolingual language model. However, in this case, monolingual languages ($L1$ and $L2$) are used in the training data and code-switching language is used in the test data. Since both monolingual languages are not the same size, hence, words are concatenated to make both languages of equal size. The classical RNN language model is then trained. The equation of RNN language model is given below:

$$\theta = argmax_\theta\ \mathbb{E}_{(x,y) \in D}\ [P_\theta(y|x)] \ldots \ldots \ldots \ldots (2)$$

However, RNN model is trained on minibatches of training data. During training, the training dataset is divided in minibatches then the equation (2) becomes:

$$\theta = argmax_\theta\ \mathbb{E}_{B \subset D}\ [\textstyle\sum_{(x,y) \in B} P_\theta(y|x)] \ldots \ldots (3)$$

In order to train RNN model on only monolingual data, one batch from first language is taken and the probability of next word is computed. The first batch training generates output predictions for next word and hidden vector. Then, a batch of data from second language is given as input, along with the hidden vector generated from previous step and then, the output prediction is computed. These steps are repeated during the training by taking alternative batches from both monolingual languages and the output predictions are computed. In short, the hidden vector from previous steps and alternate batches from monolingual languages are subjected as input to train the RNN model. Using this method, the RNN model learns weights of both monolingual training data. A block diagram of the whole process is given in figure 1. It can be observed that during the training, both monolingual languages are used in the input process to the neural network block to compute output predictions. The output of the neural network is computed with cross-entropy objective function. The mathematical equation for cross-entropy is given below:

$$H(p,q) = -\textstyle\sum pi \log qi\ \ldots \ldots \ldots \ldots \ldots \ldots (4)$$

Where $p_i \in \{y, 1-y\}$ and $q_i \in \{\acute{y}, 1-\acute{y}\}$

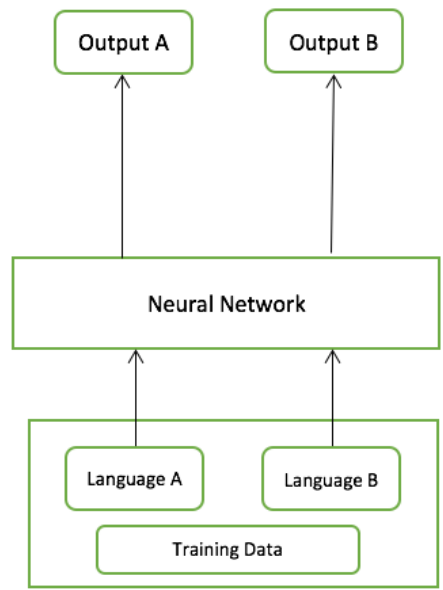

Figure: 1 Training of the RNN Model

### 2.1 Output Embeddings

The output embedding from the RNN model is the output obtained after RNN layer and it is denoted as W. First step is to compute $W1$ by using language $L1$ in training and then compute $W2$ using language $L2$. Technically, these processes can be represented as:

$W1$ = output embedding ($L1$)
$W2$ = output embedding ($L2$).

Next step is to compute mean square error (MSE) between $W1$ and $W2$ to minimize the distance between embedding

space of language (*L1*) and language (*L2*). The mathematical equation of MSE is follows:

$$L(W_1, W_2) = \frac{1}{N} \sum_{i=0}^{N}(W_1 - W_2)^2 \quad \ldots\ldots\ldots\ldots(5)$$

MSE acts like a regularization between *L1* and *L2*. The MSE is computed for all the batches of training data. The language model is then trained with joint losses of cross-entropy and the MSE and then, the weights are updated using backpropagation. The total loss then becomes:

$$Loss_{total} = H(p, q) + L(W_1, W_2) \quad \ldots\ldots\ldots\ldots(6)$$

The whole process is depicted in figure 2. In this case, the two losses are computed instead of one loss as compared to previous method.

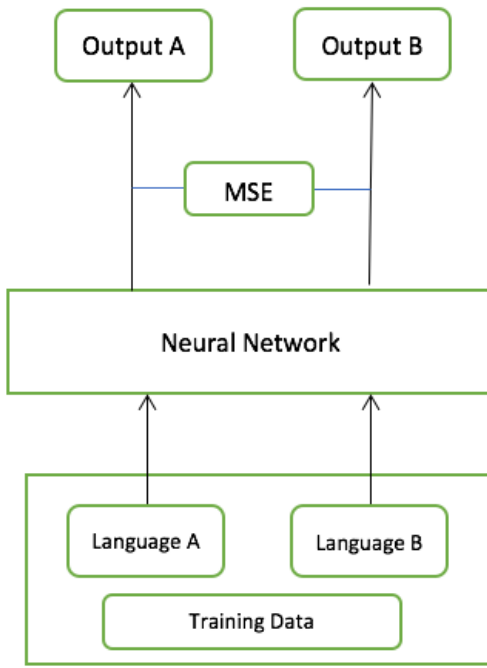

Figure:2 RNN Model training with MSE loss

## 3. EXPERIMENTS

### 3.1 Dataset

We used dataset from [11]. The dataset consists of monolingual training data of Spanish and English languages. The test data contains mixed English and Spanish. Here English is matrix language and Spanish is the embedding language. The monolingual training data from [11] is split into the English subset and the Spanish subset. The English subset consists of 4502624 words while Spanish subset consists of 3940333 words. The code-switch test data consists of 64,354 mix words. For the alternate batch training we concatenate Spanish words to make both languages of equal size. The whole dataset is converted into a vocabulary dictionary which represents words as keys and embeddings as values. We used word embedding as input to the neural network. Each word embedding shows a unique representation in high dimensional space.

### 3.2 Implementation

We trained a word level language model [15] using Pytorch framework. This model predicts next word given a sequence of previous words. The architecture consists of input embedding layer, one lstm layer and one output layer. The dimension of embedding layer is 300, lstm layer is 650 and output layer dimension is vocabulary size. The batch size of 40 and a dropout of 0.3 is used. The initial learning rate was 20 and after each epoch we divide it by 2. The language model is trained for 20 epochs and increasing above 20 does not improve the perplexity of the language model. All other hyperparameters were set according to [15].

### 3.3 Results

In this paper, the performance of code-switching language model using perplexity is measured. Perplexity is defined as the normalized inverse probability of data.

In the start, only monolingual data is used in training and then, the results are validated with code-switching test data. In the first experiment, only the monolingual Spanish data is used in the training and then, the results on code-switch test are computed as shown in first row from the Table I. The perplexity of the code-switching test data is very high (299.63). Similarly, only monolingual English data is used in the training and results are evaluated on code-switch test data as shown in second row of Table I. Then, both the monolingual languages are used in training data. We alternatively took batches from both monolingual data and compute perplexity. Using alternate batches improves the results as compared to above results. Row 4 of Table I show perplexity improvement as compared to only monolingual training data. Finally, the MSE is minimized between output embeddings and the results are computed. The last row of Table I shows the consistency in getting the improvement from 299.63 to 80.38. It can be observed that using alternate batches and MSE between output embeddings improve overall perplexity results.

All these results are also shown in a bar chart in figure 3 for comparison purpose.

| Training data | Perplexity |
|---|---|
| Spanish data only | 299.63 |
| English data only | 244.51 |
| Spanish + English data (*) | 83.10 |
| MSE (+) | 80.38 |

Table I: Results

* Trained with alternate batches from Spanish and English
+ MSE between output embeddings

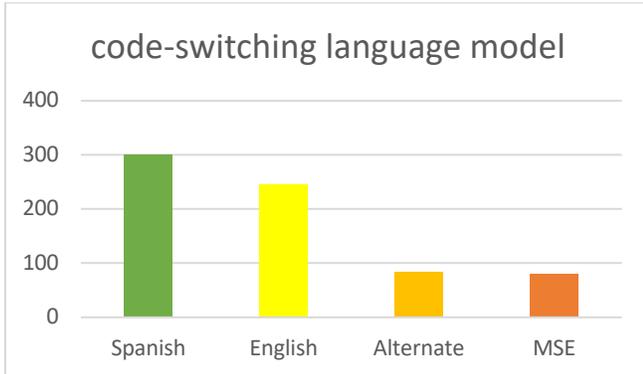

Figure 3:

## 4. CONCLUSION

In this paper, the performance of code-switching language model is evaluated. The monolingual data is used efficiently in training of the RNN model to compute the perplexity of code-switching data during evaluation. By using alternate batches from monolingual data in training, the perplexity of the code-switching test data is reduced. Moreover, this technique consistently improves results using MSE between output embeddings. From these results, it is concluded that alternate batches and MSE between output embeddings can improve the results using monolingual languages in training on code-switching test data. In future, the code-switching language model will be integrated with automatic speech recognition (ASR) to improve a code-switched ASR system.

## 5. REFERENCES

[1] T. Mikolov, M. Karafiát, L. Burget, J. H. Cernocký and S. Khudanpur, "Recurrent neural network based language model", International Speech Communication Association (INTERSPEECH), Japan, 2010.

[2] P.E James, M.H Kit, C.A. Vaithilingam and A. T. W. Chiat, "Recurrent neural network based speech recognition using MATLAB", International Journal of Intelligent Enterprise, Malaysia, 2020.

[3] M. V. Mahoney, "Text Compression as a Test for Artificial Intelligence", Conference on AAAI/IAAI, USA, 1999.

[4] H. Adel, N. T. Vu, F. Kraus, T. Schlippe, H. Li and T. Schultz, "Recurrent neural network language modeling for code switching conversational speech", International conference on Acoustics, Speech and Signal Processing (ICASSP), Canada, 2013.

[5] K. Bali, A. W. Black, R. K. Mehta, T. Solorio, V. Soto S. Sitaram and E. Yilmaz, https://www.microsoft.com/en-us/research/event/workshop-on-speech-technologies-for-code-switching-2020/, China, 2020

[6] S. Garg, T. Parekh and P. Jyothi, "Dual language models for code switched speech recognition", arXiv preprint arXiv:1711.01048, 2017.

[7] G. I Winata, A. Madotto, C. Wu and P. Fung, "Learn to code-switch: Data augmentation using copy mechanism on language modeling", arXiv preprint arXiv:1810.10254, 2018.

[8] E. Yilmaz, H. Heuvel and D. Leeuwen, "Acoustic and textual data augmentation for improved ASR of code-switching speech", arXiv preprint arXiv:1807.10945, 2018.

[9] S. Chuang, T. Sung and H. Lee, "Training a code-switching language model with monolingual data", International conference on Acoustics, Speech and Signal Processing (ICASSP), Spain, 2020.

[10] Y. Khassanov, H. Xu, V.T. Pham, Z. Zeng, E. S. Chng, C. Ni and B. Ma, "Constrained Output Embeddings for End-to-End Code-Switching Speech Recognition with Only Monolingual Data", International Speech Communication Association (INTERSPEECH), Austria, 2019.

[11] H. Gonen and Y. Goldberg, "Language modeling for code-switching: Evaluation, integration of monolingual data, and discriminative training", arXiv preprint arXiv:1810.11895, 2018

[12] A. Baheti, S. Sitaram, M. Choudhury and K. Bali, "Curriculum Design for Code-switching: Experiments with Language Identification and Language Modeling with Deep Neural Networks", International conference on Natural Language Processing (ICON), India, 2017.

[13] G.I. Winata, S. Cahyawijaya, Z. Lin, Z. Liu, P. Xu and P. Fung, "Meta-Transfer learning for Code-Switched Speech Recognition", arXiv preprint arXiv:2004.14228, 2020

[14] G. R. Madhumani, S. Shah, B. Abraham, V. Joshi and S. Sitaram, "Learning not to Discriminate: Task Agnostic Learning for Improving Monolingual and Code-switched Speech Recognition", arXiv preprint arXiv:2006.05257, 2020

[15] "Word level language model", https://github.com/pytorch/examples/tree/master/word_language_model